\documentclass{article}
\usepackage{new_template/spconf,amsmath,graphicx,hyperref}
\usepackage{algorithmic}
\usepackage{graphicx}
\usepackage{textcomp}
\usepackage{xcolor}
\usepackage{booktabs}
\usepackage{multirow}
\usepackage{subcaption}
\usepackage{amssymb}

\title{One-Shot Adaptive Segmentation For Scientific Images}
\name{Tejaswi V. Panchagnula, Allison M. Davis and Fengqing Zhu\thanks{This work is supported by the Office of Naval Research under the Multidisciplinary University Research Initiatives (MURI) Grant Award N000142412545.}}
\address{Elmore Family School of Electrical and Computer Engineering\\
Purdue University\\
West Lafayette, Indiana, USA 47907\\
\texttt{\{tpanchag, davi1987, zhu0\}@purdue.edu}}
\begin{document}
%
\maketitle

\begin{abstract}
Scientific image segmentation methods rely on extensive annotation and task-specific training, limiting adaptation across imaging modalities and experimental conditions. We present a training-free, one-shot framework that specializes vision foundation models using a single annotated reference image. The framework combines DINOv3 representations with background-adaptive feature orthogonalization to suppress artifact-related feature directions, after which cosine similarity localizes candidate regions for SAM segmentation. We evaluate the framework on red-blood-cell microscopy, structured-illumination pool boiling, and chest radiography. Relative to the strongest baseline, the proposed method improves mean IoU by $5.91\%$ and $78.62\%$ on the microscopy and pool-boiling datasets, respectively, while achieving comparable performance on chest radiographs. These results demonstrate that one-shot reference conditioning can adapt general-purpose vision models to specialized scientific segmentation tasks.
\end{abstract}


%
\section{Introduction}

\begin{figure*}[htbp] 
    \centering
    \includegraphics[width=\textwidth]{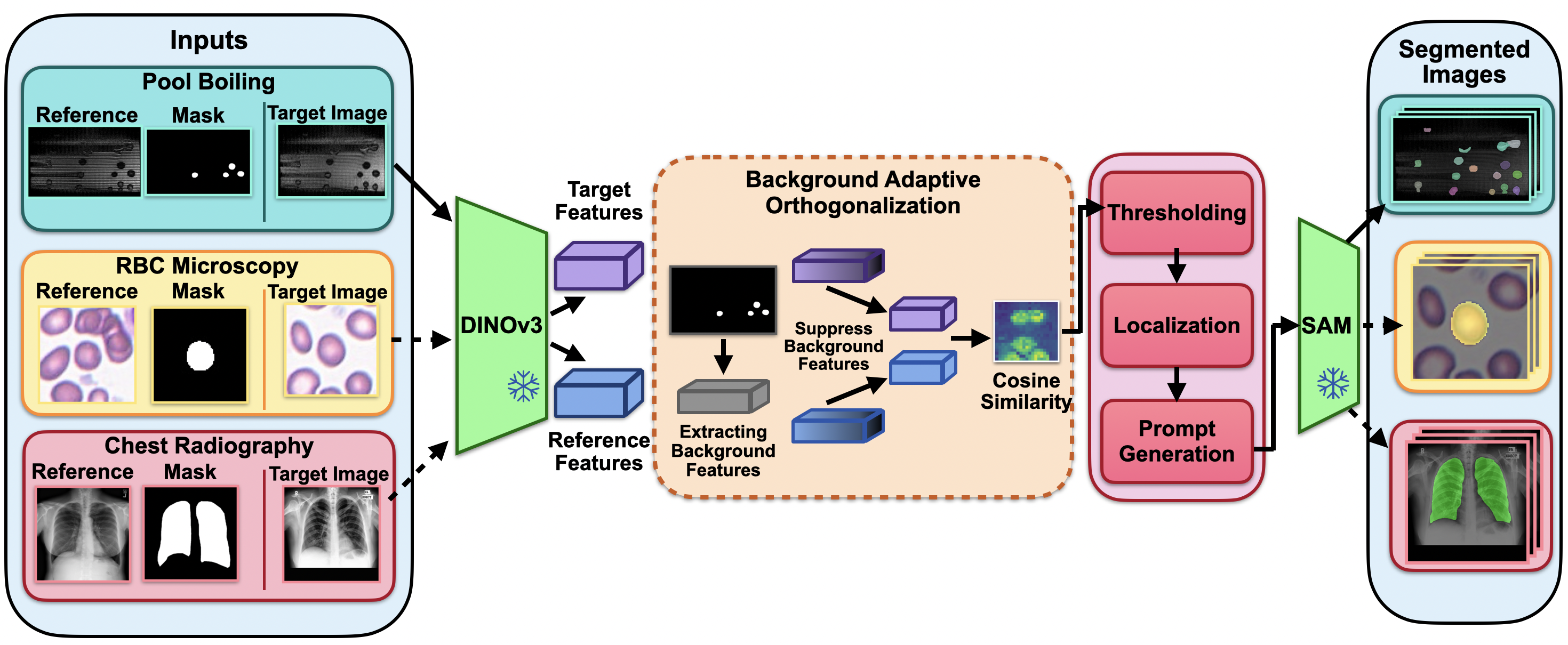}
    \caption{\textbf{Generalist-to-specialist one-shot segmentation across scientific imaging domains.} A single annotated reference defines each target concept, and the same frozen DINOv3--SAM framework localizes and segments corresponding structures without task-specific training.}
    \label{fig:fig1}
\end{figure*}

Scientific image segmentation enables quantitative analysis in applications such as multiphase flows, blood-cell microscopy, and radiography \cite{do2013image,elsafty20241,zhou2021detection}. Unlike many natural images, scientific images often contain domain-specific structures alongside acquisition-dependent artifacts and background variations, making robust segmentation challenging. Deep-learning-based approaches, including convolutional neural networks (CNNs) and vision transformers (ViTs), have shown success in segmenting scientific images. However, training a supervised model for each experimental setting and segmentation task requires extensive expert annotation.

Vision foundation models offer an alternative by transferring representations learned from large-scale pretraining to new tasks without requiring task-specific training. Among these models, self-supervised feature encoders and promptable segmentation models provide complementary capabilities for feature localization and segmentation. Segment Anything (SAM) \cite{kirillov2023segment} generates detailed segmentation masks from spatial prompts, while self-supervised models such as DINOv3 \cite{caron2021emerging,simeoni2025dinov3} provide patch-level representations that support semantic correspondence across images. However, their direct application to scientific images remains unreliable, motivating a generalist-to-specialist framework that adapts to a specific task from one annotated example.

Training-free one-shot segmentation transfers target information from a single annotated reference to localize corresponding structures in unlabeled images. Existing approaches include GF-SAM, which uses foreground and background reference features to guide SAM \cite{zhang2024bridge}, and INSID3, which mitigates positional bias in DINOv3 representations through orthogonal feature projection \cite{cuttano2026insid3}. However, these approaches do not adapt their feature correction to the background and imaging artifacts observed in the annotated reference.

We address this limitation with a training-free, background-adaptive framework for one-shot scientific-image segmentation, shown in Fig.~\ref{fig:fig1}. Our method uses a single annotated reference image and its corresponding mask to localize target structures and produce fine-grained segmentation masks. We evaluate our framework across pool-boiling imaging, red blood cell microscopy, and chest radiography, representing distinct scientific-imaging modalities and conditions.

Our main contributions in this paper are as follows:
\begin{itemize}
    \setlength{\itemsep}{0pt}
    \setlength{\parskip}{0pt}
    \setlength{\parsep}{0pt}
    \setlength{\topsep}{0pt}
    \item We present a training-free, one-shot framework that adapts frozen foundation vision models to scientific images using only a single annotated reference.
    \item We introduce a background-adaptive orthogonal projection that uses the annotated reference background to suppress background structures and imaging artifacts.
    \item We evaluate the framework across three scientific datasets, demonstrating its applicability to distinct structures, artifacts, and modalities. 
\end{itemize}

\section{Methodology}
\label{sec:methodology}


\begin{figure*}[htbp] 
    \centering
    \includegraphics[width=\textwidth]{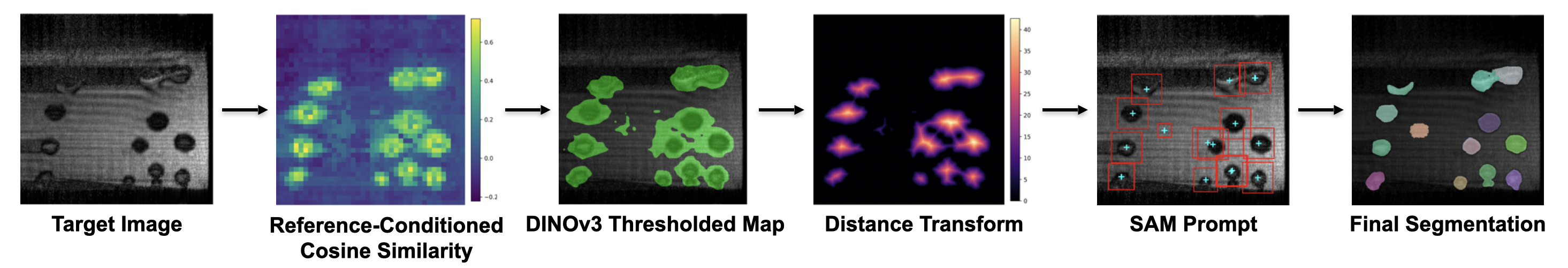}
    \caption{\textbf{End-to-end one-shot segmentation framework.}
    DINOv3 features from the target image are compared with a foreground prototype obtained from the annotated reference. The resulting similarity map is thresholded to localize candidate regions. A distance transform identifies interior point prompts, while connected regions provide bounding-box prompts for SAM2, which produces the final segmentation masks.}
    \label{fig:methods}
\end{figure*}

The proposed framework combines the semantic representations of DINOv3 with the prompt-based segmentation capabilities of SAM. A single annotated reference example, consisting of an image and its corresponding mask, defines the target structures to be localized in subsequent target images. This formulation enables adaptation to a specific segmentation objective without additional training. A qualitative visualization of the complete framework, including feature matching, candidate-region localization, prompt generation, and final segmentation, is shown in Fig.~\ref{fig:methods}.

\subsection{Feature Extraction with DINOv3}

The proposed method first passes the annotated reference frame and subsequent target frames through a frozen pretrained DINOv3 model \cite{simeoni2025dinov3}. DINOv3 is well suited to this setting because its self-supervised training produces patch-level representations which are able to group semantically similar regions without predefined class labels. For each image, we extract the normalized patch tokens from the final attention layer, giving us the feature vector $\mathbf{F} \in \mathbb{R}^{H' \times W' \times D}$, where \emph{D} is the embedding dimension and $H'$ and $W'$ are the dimensions of the patch-wise grid produced by the ViT. The annotated reference image can therefore define a new target class from its feature representation.

\subsection{Background-Adaptive Feature Orthogonalization}

Scientific images can contain acquisition-specific artifacts, including glare, occluded objects, and motion blur in the raw image. During post-processing, these artifacts may not be removed completely. To reduce the effects of these components in the segmentation process, we employ a latent-space orthogonal projection. The reference mask is downsampled to match the spatial dimensions of the DINOv3 patch grid. The resulting Boolean mask partitions the reference tokens into foreground and background sets, with the selected foreground tokens forming the feature set $\mathbf{F}_{\mathrm{ref}}$.


We obtain a unit background direction, $\mathbf{v}_c$, by averaging and $L_2$-normalizing the background tokens. A raw foreground prototype, $\mathbf{p}_{\mathrm{raw}}\in\mathbb{R}^{D}$, is similarly computed by averaging the foreground tokens. For any feature vector $\mathbf{x}$, its normalized projection onto the orthogonal complement of $\mathbf{v}_c$, shown in (\ref{eq:orthogonal_projection}) as

\begin{equation}
\mathcal{P}_{\perp}(\mathbf{x}) =
\frac{
\mathbf{x}-(\mathbf{x}^{\top}\mathbf{v}_c)\mathbf{v}_c
}{
\left\|
\mathbf{x}-(\mathbf{x}^{\top}\mathbf{v}_c)\mathbf{v}_c
\right\|_2
}.
\label{eq:orthogonal_projection}
\end{equation}
The projected reference features are $\mathbf{p}=\mathcal{P}_{\perp}(\mathbf{p}_{\mathrm{raw}})$. During inference, the same operation is applied independently to every target token $\mathbf{f}_i$, producing $\mathbf{f}^{\,\prime}_i=\mathcal{P}_{\perp}(\mathbf{f}_i)$. The resulting target feature tensor is formed by collecting the projected tokens, $\mathbf{F}_{\mathrm{out}} =[\mathbf{f}'_1,\mathbf{f}'_2,\ldots,\mathbf{f}'_N]^\top \in\mathbb{R}^{N\times D}$, where $N$ is the number of image patches and $D$ is the feature dimension. The reference mask need not contain every target instance, provided that the annotated regions adequately represent the target structure.


\subsection{Localization, Prompt Generation, and Segmentation}

To localize candidate target regions, we compute cosine similarity between the projected reference prototype $\mathbf{p}$ and every projected target token in $\mathbf{F}_{\mathrm{out}}$. The resulting scores are reshaped into a two-dimensional similarity map and thresholded to produce a binary candidate map. Because patch-level similarity can create small gaps within otherwise coherent regions, a morphological hole-filling operation is applied to consolidate the detected structures. A Euclidean distance transform then assigns each foreground pixel its distance to the nearest background pixel. Local maxima of the distance map provide interior point prompts, while the enclosing bounds of each connected region provide bounding-box prompts. These prompts are supplied to SAM, which produces a fine-grained segmentation mask for the target structures represented by the annotated reference.
\section{Results}
\label{sec:results}


\begin{table*}[htbp]
    \centering
    \caption{One-shot segmentation performance across scientific imaging domains. Best results are shown in bold and second-best results are underlined.}
    \label{tab:metrics}
    \begin{tabular}{@{}lllcc@{}}
        \toprule
        \textbf{Dataset} & \textbf{Imaging Modality} & \textbf{Method} & \textbf{Mean Target IoU} & \textbf{Mean Target Dice (F1)} \\
        \midrule
        PathOlOgics (RBCs) & Microscopy & SAM2 & 0.8670 & 0.9201 \\
        PathOlOgics (RBCs) & Microscopy & GF-SAM & \underline{0.8720} & \underline{0.9297} \\
        PathOlOgics (RBCs) & Microscopy & INSID3 & 0.7670 & 0.8644 \\
        PathOlOgics (RBCs) & Microscopy & \textbf{Ours} & \textbf{0.9235} & \textbf{0.9591} \\
        \midrule
        Pool Boiling & SI Reconstructed & SAM2 & 0.0600 & 0.1127 \\
        Pool Boiling & SI Reconstructed & GF-SAM & 0.1646 & 0.2761 \\
        Pool Boiling & SI Reconstructed & INSID3 & \underline{0.4243} & \underline{0.5950} \\
        Pool Boiling & SI Reconstructed & \textbf{Ours} & \textbf{0.7579} & \textbf{0.8612} \\
        \midrule
        Montgomery Chest X-Ray & Chest Radiograph & SAM2 & \textbf{0.3962} & \underline{0.5608} \\
        Montgomery Chest X-Ray & Chest Radiograph & GF-SAM & 0.3805 & 0.5434 \\
        Montgomery Chest X-Ray & Chest Radiograph & INSID3 & 0.3044 & 0.4637 \\
        Montgomery Chest X-Ray & Chest Radiograph & \textbf{Ours} & \underline{0.3960} & \textbf{0.5610} \\

        \bottomrule
    \end{tabular}
\end{table*}



We evaluate the proposed method across red-blood-cell microscopy, structured-illumination pool boiling, and chest radiography. These datasets represent distinct imaging modalities, target structures, and background characteristics, allowing us to assess the adaptability of the same framework under different scientific imaging conditions.

\subsection{Datasets}
\noindent\textbf{Red Blood Cell Microscopy.} We first use the 100k-RBC-PathOlOgics dataset \cite{elmanna2024deep}, which contains 240,790 segmented RBC images across nine classes with varying cell structures. We use a separate annotated reference for each class to assess the framework's ability to localize and delineate small targets with subtle class-specific differences.

\noindent\textbf{Structured Illumination Pool Boiling.} We evaluate 20 pool-boiling frames recorded using structured illumination (SI) and reconstructed for optical sectioning. Two different reconstruction methods are evaluated: the Sequence Hilbert Transform (SHT) reconstruction \cite{zhou2015double} and the root-mean-square (RMS) reconstruction \cite{kristensson2012structured,neil1997method}. Residual illumination patterns, spatially varying contrast, and closely spaced features make accurate foreground localization particularly challenging. For each reconstruction method, the reconstructed frames serve as target images for segmentation. 

 \noindent\textbf{Chest Radiography.} Finally, we evaluate one-shot segmentation on the Montgomery County chest X-ray dataset \cite{jaeger2012detecting}, which contains 138 frontal radiographs with corresponding lung masks, including 58 tuberculosis-positive cases. This dataset tests adaptation to projection images with overlapping anatomy and low-contrast boundaries.

\subsection{Segmentation Results}

\begin{figure}[t]
    \centering
    \includegraphics[width=\columnwidth]
    {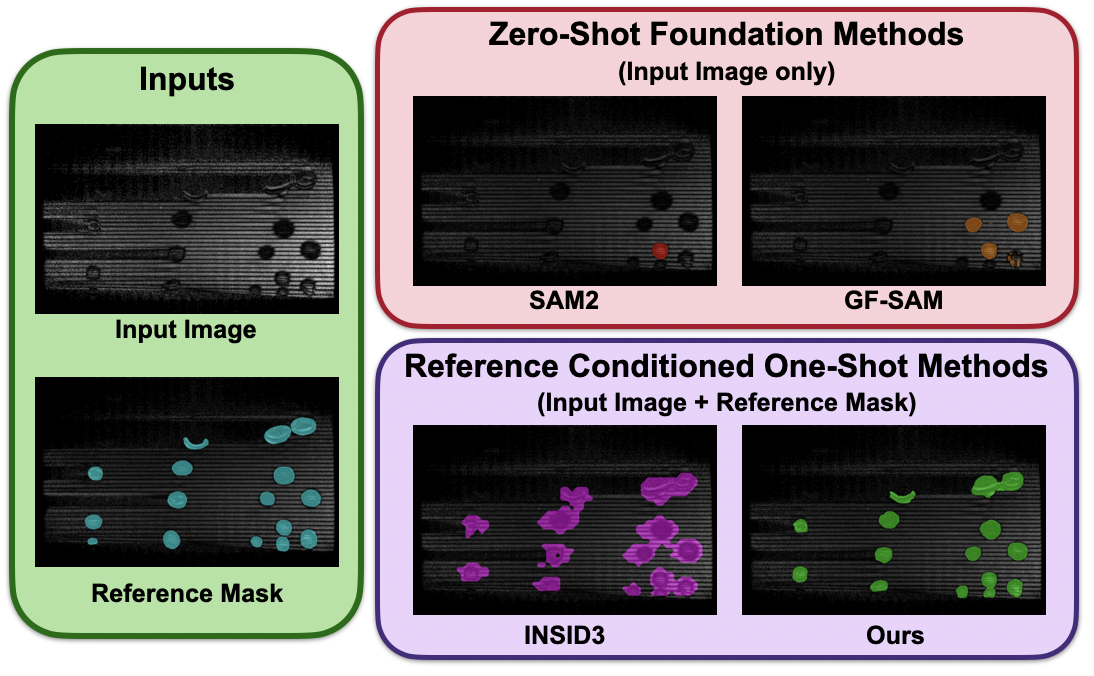}
    \caption{\textbf{Qualitative segmentation comparison} on structured-illumination pool-boiling data. The proposed method recovers more of the required annotated features than SAM2 and GF-SAM while accurately reducing the enlarged and merged regions produced by INSID3.}
    \label{fig:qualitative_comparison}
\end{figure}

We evaluate segmentation performance using mean intersection over union (IoU) and Dice score. Table~\ref{tab:metrics} compares the proposed method with SAM2, GF-SAM, and INSID3. On RBC microscopy, the proposed method exceeds GF-SAM, the strongest baseline, by 5.15 percentage points in mean IoU and 2.94 percentage points in Dice. The largest gains occur on pool-boiling images, where the proposed method exceeds INSID3 by 33.36 percentage points in mean IoU and 26.62 percentage points in Dice. On chest radiography, the proposed method achieves performance comparable to SAM2. Fig.~\ref{fig:qualitative_comparison} shows that SAM2 and GF-SAM recover only a subset of the bubbles in the pool-boiling target image, while INSID3 produces enlarged and merged masks. The proposed method recovers more instances while better preserving their shapes and boundaries. These results indicate that background-adaptive feature processing is particularly effective when scientific imaging artifacts dominate the learned feature representations.

\subsection{Background Response Suppression}

\begin{figure}[htbp]
    \centering
    \begin{subfigure}[b]{0.22\textwidth}
        \centering
        \includegraphics[width=\textwidth]{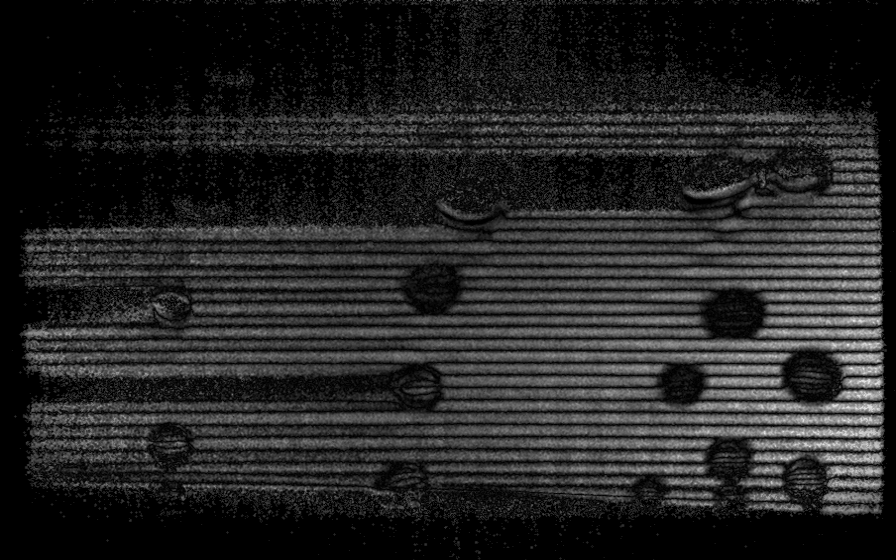}
        \caption{Target Image}
    \end{subfigure}
    \hfill
    \begin{subfigure}[b]{0.22\textwidth}
        \centering
        \includegraphics[width=\textwidth]{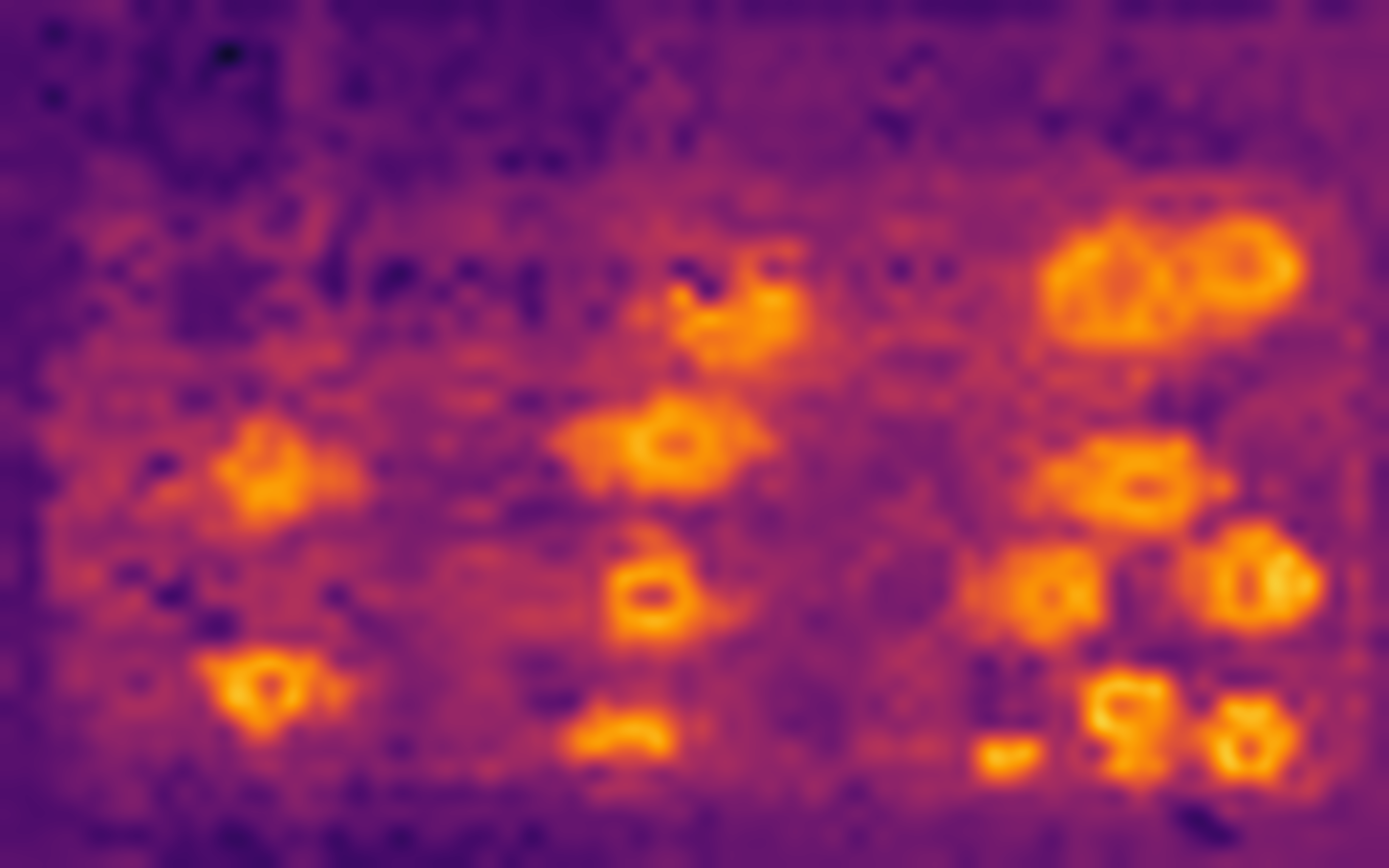}
        \caption{Unprojected DINOv3}
    \end{subfigure}
    \hfill
    \begin{subfigure}[b]{0.22\textwidth}
        \centering
        \includegraphics[width=\textwidth]{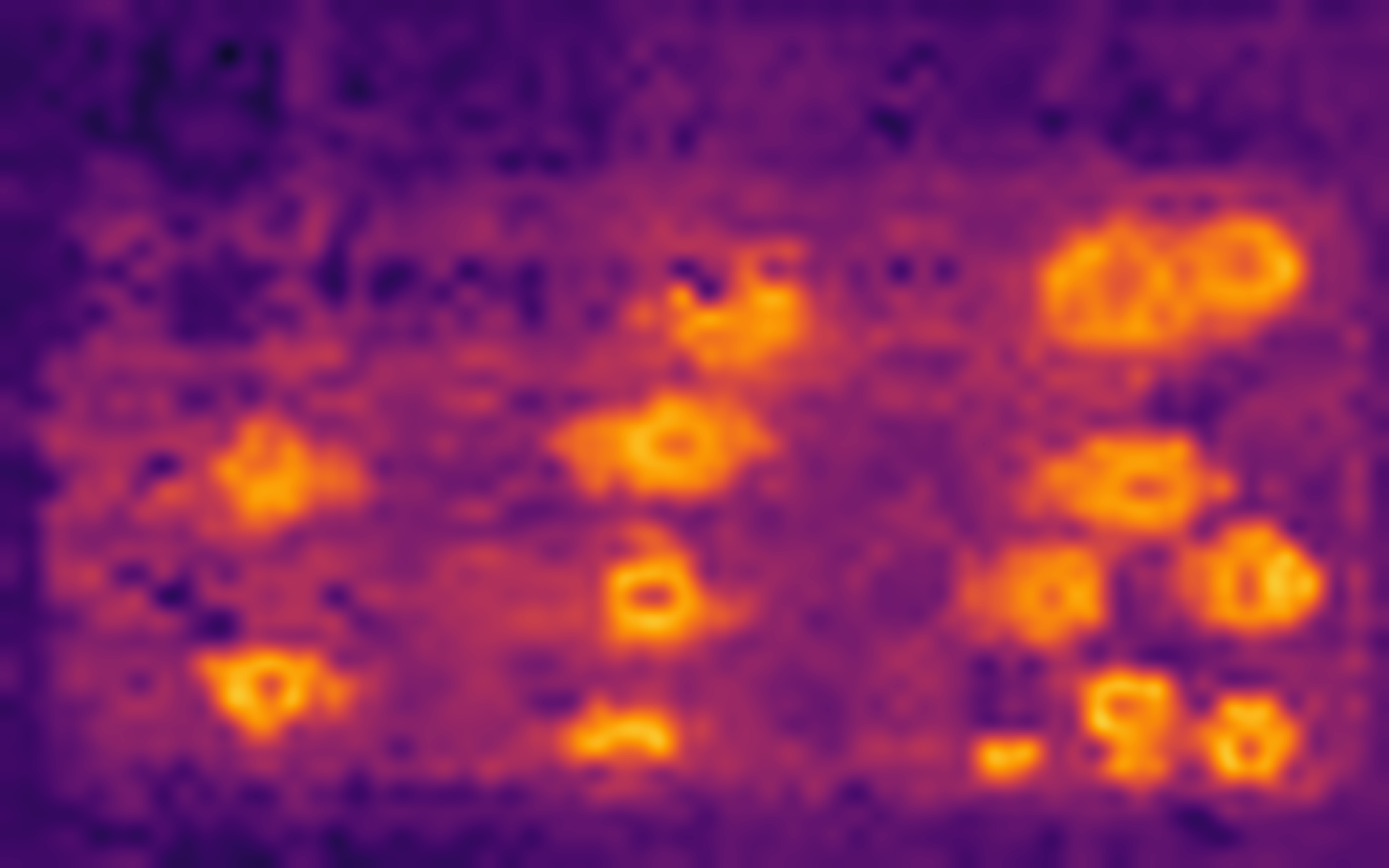}
        \caption{Gaussian Noise Projection}
    \end{subfigure}
    \hfill
    \begin{subfigure}[b]{0.22\textwidth}
        \centering
        \includegraphics[width=\textwidth]{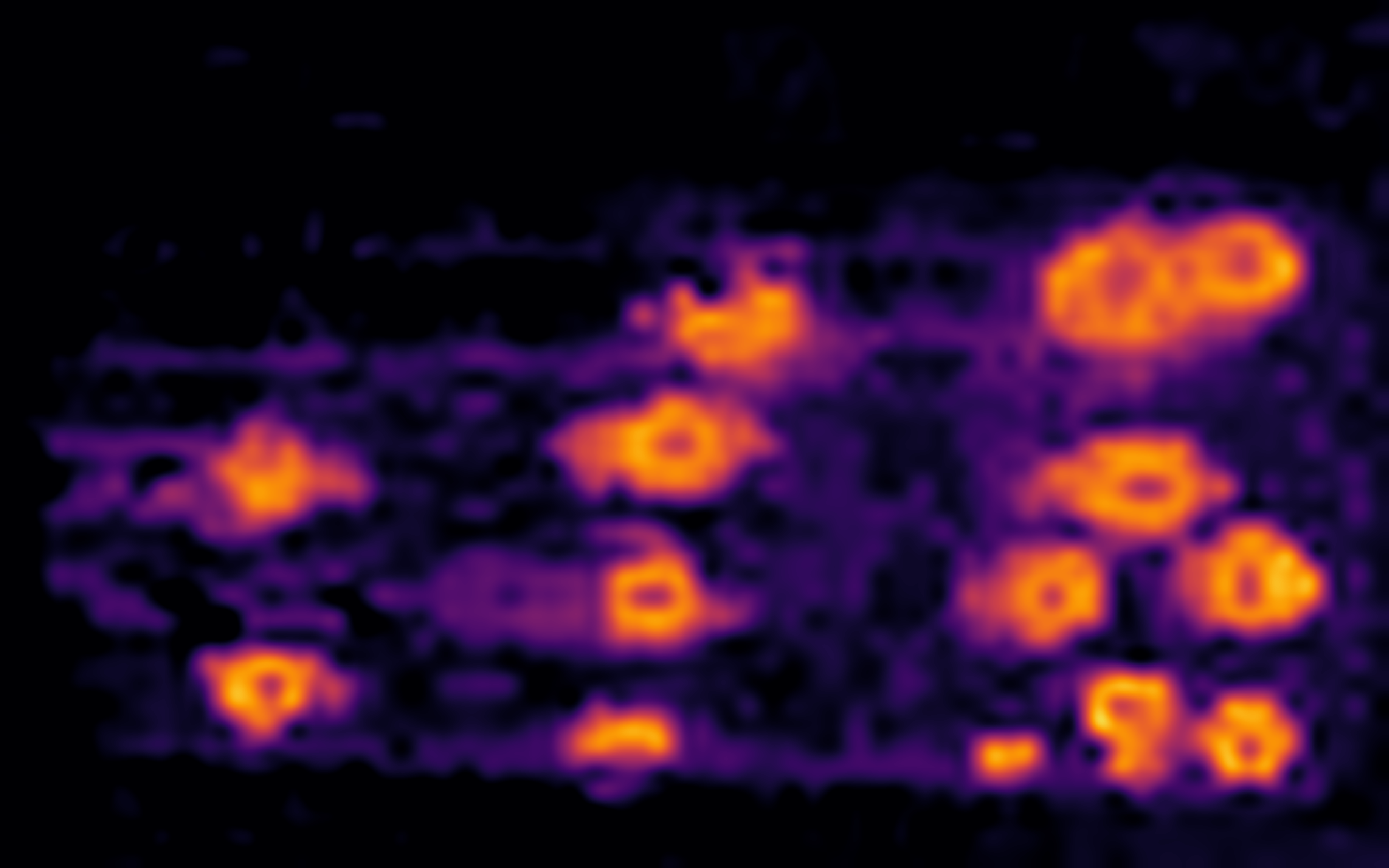}\caption{Background Adaptive}
    \end{subfigure}
    \caption{\textbf{Effect of background-adaptive feature orthogonalization} on an RMS-reconstructed pool-boiling frame. Compared with unprojected DINOv3 features and the Gaussian-noise projection, the proposed method suppresses background responses while preserving target-bubble activation.}
    \label{fig:noise_reduction}
\end{figure}


Background-adaptive feature orthogonalization substantially suppresses residual background activation in the cosine-similarity map. As shown in Fig. \ref{fig:noise_reduction}, the unprojected DINOv3 features and the Gaussian-noise projection used by INSID3 \cite{cuttano2026insid3} retain elevated responses across the structured background. The proposed method reduces these responses to near zero while preserving strong, localized activation over the target bubbles. Across 20 pool-boiling frames, mean background similarity decreases from $0.3117$ for the unprojected features and $0.2847$ for the Gaussian-noise projection to $-0.0020$ with the proposed method. The contrast-to-noise ratio correspondingly increases from $2.74$ and $2.64$ to $2.97$. These results indicate that a background direction estimated from the annotated reference provides stronger feature-space separation than a generic Gaussian-noise reference.


\section{Conclusion}
\label{sec:conclusion}

While scientific image segmentation is essential for quantitative analysis, existing approaches often require extensive expert annotation or perform unreliably under specialized imaging conditions. To address these limitations, we present a training-free, one-shot framework that adapts frozen vision foundation models for scientific image segmentation. Experiments on red blood cell microscopy, structured-illumination pool boiling, and chest radiography demonstrate the adaptability of the framework to distinct imaging modalities. On pool-boiling data, the proposed method exceeds the strongest baseline by 33.36 percentage points in mean IoU while substantially suppressing background responses in feature space. By reducing reliance on large, expert-annotated datasets, this background-adaptive framework can make specialized segmentation more accessible across scientific imaging domains.


\newpage
\bibliographystyle{new_template/IEEEbib}
\bibliography{references}

\end{document}